\documentclass[letterpaper, 10 pt, conference]{ieeeconf}
\IEEEoverridecommandlockouts                              
                                                          
\usepackage{tikz}
\usetikzlibrary{tikzmark}

\usepackage{cite}

\usepackage{amsmath}

   \usepackage{graphicx}

\usepackage[T1]{fontenc}

\usepackage{array}

\usepackage{amsmath}
\usepackage{mathtools}
\usepackage{amssymb}
\usepackage{breqn}

  \usepackage[caption=false,font=normalsize,labelfont=sf,textfont=sf]{subfig}
\usepackage{fixltx2e}

\usepackage{stfloats}
\usepackage{url}

\usepackage{algorithm}
\usepackage{algpseudocode}
\usepackage{multirow}
\usepackage{mathtools}
\usepackage{cuted}

\usepackage[utf8]{inputenc}
\usepackage[T1]{fontenc}

\begin{document}
%
\title{\LARGE \bf Adaptive Cost-Map-based Path Planning in Unknown Environments with Movable Obstacles}

\author{
Liviu-Mihai Stan$^{1,3}$, Ranulfo Bezerra$^{1,2}$, Shotaro Kojima$^{1,2}$, Tsige Tadesse Alemayoh$^{1,2}$, Satoshi Tadokoro$^{1,2}$ \\
Masashi Konyo$^{1}$, Kazunori Ohno$^{1,2}$
\thanks{*This research was performed by the commissioned research fund provided by F-REI (JPFR23010101).}
\thanks{$^{1}$Graduate School of Information Sciences, Tohoku University, Japan.
}
\thanks{$^{2}$Tough Cyberphysical AI Research Center, Tohoku University, Japan.
}
\thanks{$^{3}$Sorbonne University, Paris, France.
}
\thanks{\tt\small bezerra.ranulfo@tr.is.tohoku.ac.jp}
\thanks{\tt\small liviu.stan99@gmail.com}
}


%


\maketitle

\begin{abstract}
Reliable navigation in disaster-response and other unstructured indoor settings requires robots not only to avoid obstacles but also to recognise when those obstacles can be pushed aside. We present an adaptive, \mbox{LiDAR-only} path-planning framework that embeds this capability into the ROS\,2 \textit{Nav2} stack without any vision pipeline or pre-training. A new \textit{Movable Obstacles Layer} labels all LiDAR returns missing from a prior static map as tentatively movable and assigns a reduced traversal cost. A companion \textit{Slow-Pose Progress Checker} monitors the ratio of commanded to actual velocity; when the robot slows appreciably, the local cost is raised from \emph{light} to \emph{heavy}, and on a stall to \emph{lethal}, prompting the global planner to back out and re-route. Gazebo evaluations on a Scout Mini, spanning isolated objects and cluttered corridors, show higher goal-reach rates and fewer deadlocks than a no-layer baseline, with traversal times broadly comparable. Because the method relies only on planar scans and CPU-level computation, it suits resource-constrained SAR robots and integrates into heterogeneous platforms with minimal engineering. Overall, the results indicate that interaction-aware cost maps are a lightweight, ROS\,2-native extension for navigating among potentially movable obstacles in unstructured settings. The full implementation will be released as open source at \url{https://costmap-namo.github.io}.
\end{abstract}
%
\IEEEpeerreviewmaketitle

\section{Introduction}

Reliable autonomous navigation is a cornerstone capability for disaster response robotics. In collapsed buildings, smoke filled hospital corridors, or earthquake damaged industrial sites, human access may be hazardous or impossible. Search and rescue (SAR) robots must reach victims, deliver supplies, and explore confined spaces without assistance. The ability to navigate such environments reliably is essential for saving lives, minimizing rescue times, and reducing risks to human responders.

Autonomous robots are increasingly expected to operate in cluttered, \emph{unstructured} indoor environments, such as collapsed corridors or makeshift passages in disaster zones. Researchers refer to this challenge as \emph{Navigation Among Movable Obstacles} (NAMO), where the robot must plan paths that may involve interacting with and repositioning obstacles to reach its goal. In these SAR scenarios, the robot’s survival critical task is to reach targets despite debris, overturned furniture, or equipment that \emph{might} be movable but whose properties are unknown. Planners that assume static obstacles, or depend on camera based semantic maps easily degraded by dust and poor lighting, halt as soon as the nominal path is blocked. Approaches that attempt to \emph{move} obstacles often require rich RGB-D perception or offline learning to decide \emph{which} objects can be pushed \cite{zherdev_swipebot_2023}\cite{yao_local_namo_rl}.

However, such perception heavy methods are fragile in SAR, where visibility is reduced by dust or smoke and where time and power budgets preclude GPU based inference. Vision pipelines may fail outright or yield unreliable classifications, leaving the robot unable to progress when faced with unknown obstacles.

\begin{figure}[t!]
\centering
\includegraphics[width=0.48\textwidth]{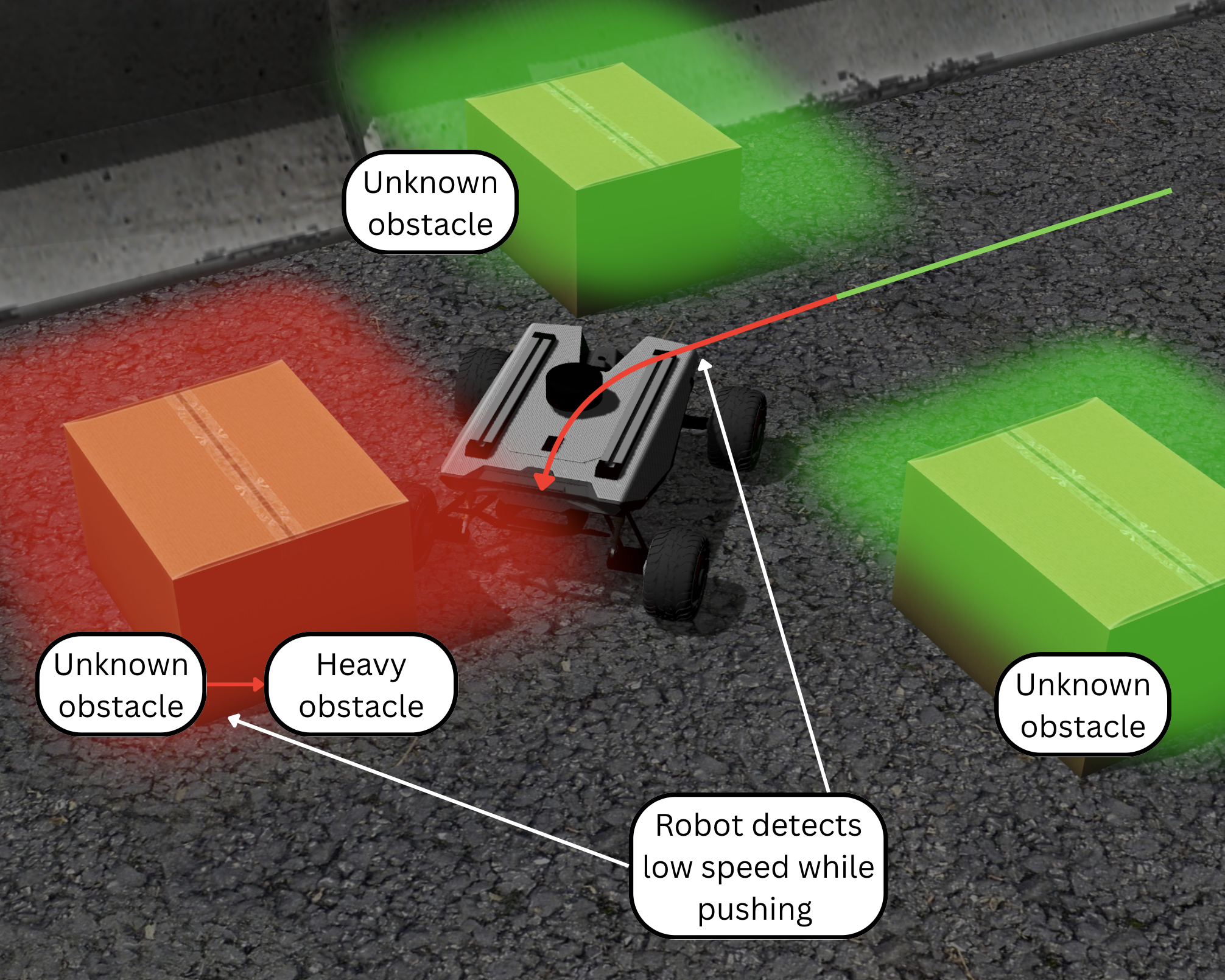}
\caption{Scout Mini in a corridor world. The robot pushes an obstacle, detects reduced speed, and labels it as heavy.}
\label{Fig:scout}
\end{figure}

The problem addressed here is how to enable a resource constrained SAR robot to determine, in real time and without vision or prior semantics, whether an obstacle can be moved, and to adjust its navigation plan accordingly. This must be done entirely within a standard ROS\,2 navigation stack for full compatibility and minimal computational overhead \cite{macenski_marathon_2020}.

We propose a LiDAR only, adaptive cost map approach implemented as two lightweight Nav2 plugins. Real time LiDAR scans are compared with a static map; new returns are labelled \emph{movable} in a \emph{Movable Obstacles Layer}. During execution, a custom \emph{Slow Pose Progress Checker} monitors the ratio between commanded and actual velocity: if motion slows, the closest obstacle is escalated from \emph{light} to \emph{heavy}, and if stalled, to \emph{lethal}, prompting replanning. Both plugins run at sensor rate, require no vision, and integrate seamlessly with existing planners. Gazebo trials from single box hallways to double row clutters show high success rates where baseline static obstacle planners fail.

\medskip
\noindent \textit{Main contributions}
\begin{itemize}
\item \textbf{ROS\,2 native, SAR oriented NAMO formulation} eliminating vision and semantic prerequisites, directly integrable with Nav2.
\item \textbf{Movable Obstacles Layer} for Nav2 costmaps tagging LiDAR detected unknown objects with \emph{adaptive} traversal costs and updating them based on progress checker data.
\item \textbf{Velocity based progress checker} sending "heavy" or "unmovable" messages to the movable obstacles layer in real time.
\end{itemize}

The following section reviews relevant literature to contextualize these contributions.

\section{Related Work}
Stilman and Kuffner’s LP1 planner first showed that Navigation Among Movable Obstacles (NAMO) can be solved exactly in the plane by decomposing free space and reconnecting components for real-time solutions with many movable objects \cite{stilman_navigation_nodate}.

Work then targeted scalability and partial observability. Wu et al. removed the need for a priori maps by incrementally building a world model from 2-D laser scans and pruning search with cost-bounding heuristics \cite{hai-ning_wu_navigation_2010}. Levihn et al. guaranteed locally optimal decisions in unknown environments using admissible bounds and dual-list pruning, scaling to tens of obstacles without losing real-time performance \cite{levihn_locally_2014}. Clingerman et al. recast cost learning as Bayesian evidence-grid inference with gamma-distributed cell costs and a lower-confidence bound driving D*-Lite replanning \cite{clingerman_dynamic_2015}.

Perception–manipulation loops added interaction to classification. Kakiuchi et al. used colour TOF sensing and force feedback on a humanoid to re-label obstacles after each push \cite{kakiuchi_working_2010}, while SwipeBot paired an FCN-ResNet RGB-D segmenter with class-dependent traversal costs updated when the motor current limit is reached \cite{zherdev_swipebot_2023}. Scholz et al. framed NAMO as a hierarchical MDP that schedules clearing actions while a physics-based RL module re-estimates mass, friction and locked-caster states from force–torque data \cite{scholz_navigation_2016}. Novin et al. learned Bayesian point-mass dynamics for hospital walkers from minimal interactions and coupled them with a hybrid MPC that selects both leg and force direction \cite{novin_dynamic_2018}. Yao et al. train an Advantage Actor–Critic policy for local NAMO that performs non-axial pushes and transfers to a Unitree Go1, enabling fast local planning among pushable objects under sensing noise \cite{yao_local_namo_rl}.

Recent planners exploit structure and priors under sensing limits. Muguira-Iturralde et al. formalize visibility-aware NAMO and propose LAMB, which interleaves visibility and manipulation subgoals to outperform NAMO-only and VAMP-only baselines in 3-D scenes with occlusions \cite{iturralde_vanamo}. He et al.’s Interactive-FAR tightly couples mapping, path-finding and manipulation via a two-level Directed Visibility Graph, re-routing in milliseconds while testing affordances through tactile feedback \cite{he_interactive-farinteractive_2024}. Ellis et al. extend online NAMO with multi-object pushing and designated storage zones, planning alternating-axis pushes and caching candidate plans to improve time efficiency while handling LP2, non-monotone dependencies \cite{ellis_multi_object_storage_2023}. 

Orthogonal work in semantic navigation, like IntelliMove, argues for hierarchical semantic topometric maps to guide planning at object/room levels, but requires richer perception than our setting.\cite{semantic_labeling}

\section{Adaptive Cost-Map-based Path Planning}

\begin{figure}[h]
\centering
\includegraphics[width=0.48\textwidth]{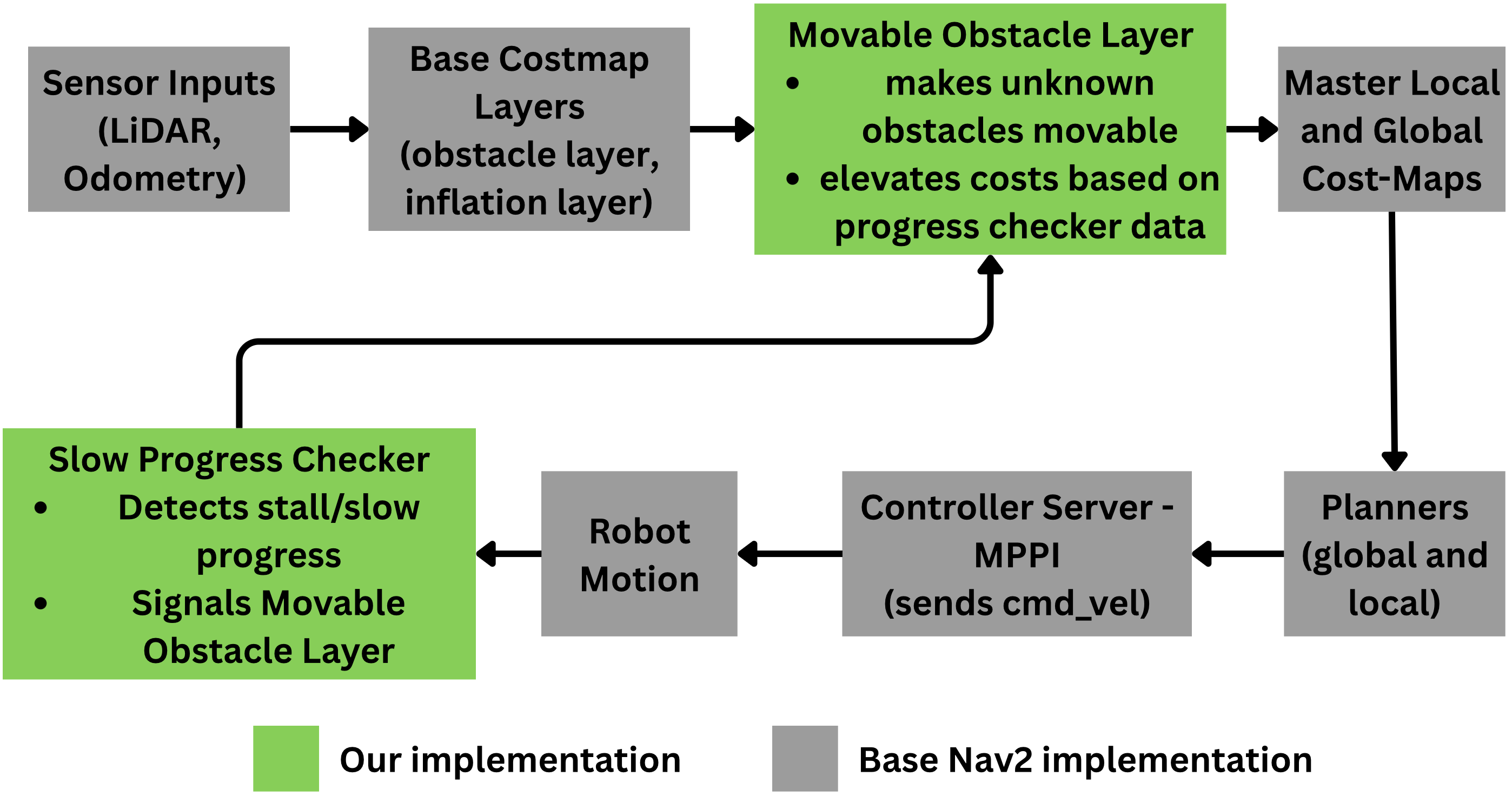}
\caption{System architecture of the proposed adaptive cost-map navigation framework. Sensor data are processed through cost-map layers, augmented by obstacle movability assessment and cost escalation, before global and local planning with MPPI control.}
\label{Fig:flowchart}
\end{figure}

\begin{figure*}[!t]  
  \centering
  \includegraphics[width=0.98\textwidth]{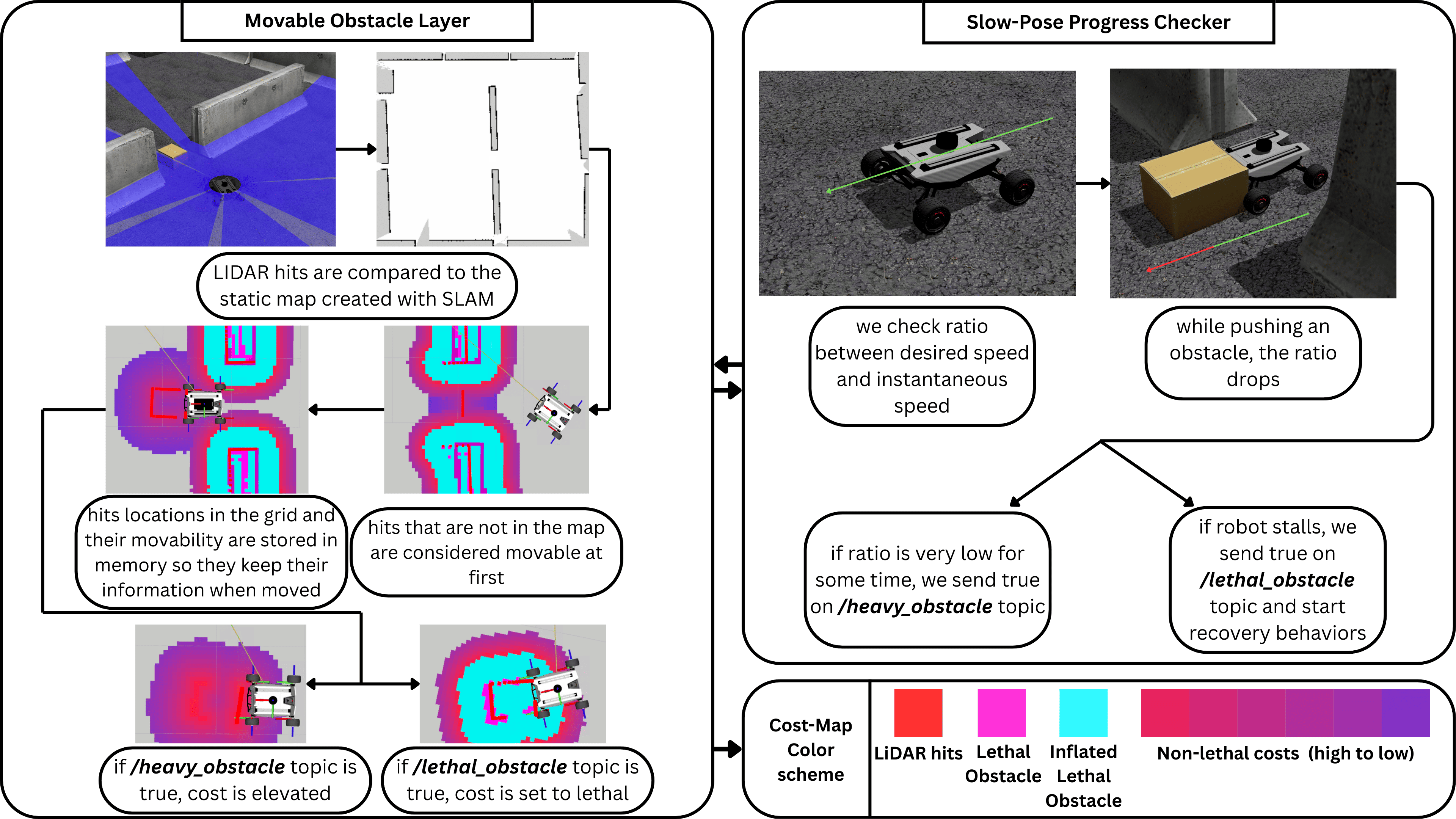}
  \caption{Detailed flowchart of the adaptive cost-map pipeline and Rviz2 Cost-Map color scheme. The Movable Obstacles Layer compares LiDAR scans to the static SLAM map to classify movable obstacles and update cost-map values, while the Slow-Pose Progress Checker monitors velocity ratios to escalate obstacle costs and trigger recovery behaviors.}
  \label{fig:project_overview}
\end{figure*}

Our system equips an AgileX Robotics Scout Mini \cite{agilex_scoutmini} robot with a LiDAR, modelled in URDF, to build a static map with SLAM Toolbox, localize via AMCL, and navigate using the ROS 2 Nav2 stack augmented by our Movable Obstacles Layer and a Slow-Pose Progress Checker (see Fig. \ref{Fig:flowchart} for the system architecture and Fig. \ref{fig:project_overview} for the detailed processing pipeline). Each incoming laser scan is compared against the static map; returns that do not coincide with known structure are tagged as movable and written into a dedicated cost-map layer with a reduced traversal cost. While the Model Predictive Path Integral (MPPI) \cite{williams_aggressive_2016} controller executes the global path, the progress checker monitors the ratio between commanded and actual velocity: if the robot slows, the nearby obstacle’s cost is escalated from light to heavy and, if a full stall is detected, to lethal, prompting the global planner to back out and re-route.


\subsection{Costmap} The navigation capabilities of the ROS 2 Navigation (Nav2) stack are fundamentally built upon the Costmap2D representation, a multi-dimensional occupancy grid where each cell encodes a traversal cost based on information from multiple layers. We extend this architecture by introducing a custom plugin, the Movable Obstacles Layer, integrated into Nav2’s layered costmap framework \cite{lu_layered_2014}.

At each update cycle, LiDAR scans, already processed by Nav2’s base Obstacle Layer, which marks all detected returns as lethal, are compared against the static SLAM map. Cells corresponding to detections not present in the static map are initially classified as movable, assigned a reduced “light” cost, and given a persistent cluster ID through a connected-component labeling process. This labeling ensures that an obstacle retains its identity and associated cost level even if temporarily occluded or partially moved. By building on the base obstacle layer rather than duplicating its functionality, our approach inherits its efficient clearing behaviour: when a detected obstacle is moved or disappears from the LiDAR view, its lethal marking is removed by the obstacle layer, and our reduced-cost cells are likewise cleared from the master costmap.

The plugin incorporates neighbourhood-based heuristics and a distance-to-wall filter to reduce false positives. Specifically, obstacles detected far enough from static walls (as determined via a precomputed distance transform) are considered movable, while those adjacent to static structures remain lethal.

Integration with the robot’s interaction feedback loop enables real-time cost escalation. Two ROS topics, \texttt{/heavy_obstacle} and \texttt{/lethal_obstacle}, allow external modules such as the Slow-Pose Progress Checker to trigger cost increases. Obstacles that impede motion without complete stalling are reclassified as heavy, increasing their cost, while obstacles that cause full stalls are escalated to lethal.

Unlike the standard inflation layer, our implementation performs weighted soft inflation for all obstacle types. Each obstacle’s base cost (light, heavy, lethal) is propagated outward within a configurable radius, decaying exponentially with distance. This yields more natural navigation: the robot prefers obstacle avoidance whenever a collision-free route exists, and initiates controlled pushing only when no such route is available or when doing so produces a better path.

This design enables the robot to adaptively and efficiently plan in cluttered environments, dynamically adjusting its traversability map based on both perception and interaction feedback.

\subsection{Path Planning and Execution}
For global path planning, our system utilizes the NavFnPlanner, a graph-based algorithm from the Nav2 stack. The global planner computes feasible paths considering the dynamically updated costmap, ensuring planned routes avoid high-cost or lethal obstacle regions.

The local path execution is managed by the Model Predictive Path Integral (MPPI) controller, selected for its superior reactive capabilities \cite{williams_aggressive_2016}. The MPPI controller, unlike traditional local planners, can perform complex maneuvers such as backward motion, enabling effective repositioning when encountering obstacles that cannot be immediately bypassed. This backward maneuvering significantly enhances local path adaptability, allowing the robot to reposition for optimal obstacle interaction, particularly beneficial after obstacles have been partially moved. This selection was validated through simulation scenarios involving complex obstacle interactions.

\subsection{Progress Checker}
To enable velocity-based obstacle cost escalation, we developed a custom ROS~2 Nav2 plugin, the \textit{Slow Pose Progress Checker}, extending the standard \texttt{PoseProgressChecker} interface. This module continuously monitors the robot’s translational speed relative to its commanded velocity, using both odometry-derived instantaneous speed and the latest \texttt{/cmd\_vel} commands.

The plugin maintains a \textit{baseline cruise speed} and a configurable \textit{drop ratio} threshold. If the measured speed drops below the configured fraction of the commanded speed for longer than a \textit{freeze window}, and the robot is not performing in-place rotation, the checker concludes that forward progress is obstructed. A short \textit{settling period} after startup and a \textit{cooldown period} between triggers prevent spurious activations.

When an obstruction is detected, the checker publishes a Boolean flag on the \texttt{/heavy\_obstacle} topic. The \textit{Movable Obstacles Layer} subscribes to this topic and escalates the cost of the nearest obstacle cluster from \textit{light} to \textit{heavy}. If further checks determine the robot is not moving at all despite attempted pushing, escalation to \textit{lethal} occurs through the \texttt{/lethal\_obstacle} topic. This direct ROS topic interface ensures low-latency coupling between velocity-based interaction feedback and the layered costmap.

By leveraging both odometry and command velocity data, the checker distinguishes between intentional slow movements (e.g., approaching a goal or turning in place) and true mobility loss. This selective escalation enables a graded response (\textit{light} to \textit{heavy} to \textit{lethal}) improving replanning decisions without prematurely abandoning pushable obstacles.


\begin{figure}[h]
\centering
\includegraphics[width=0.48\textwidth]{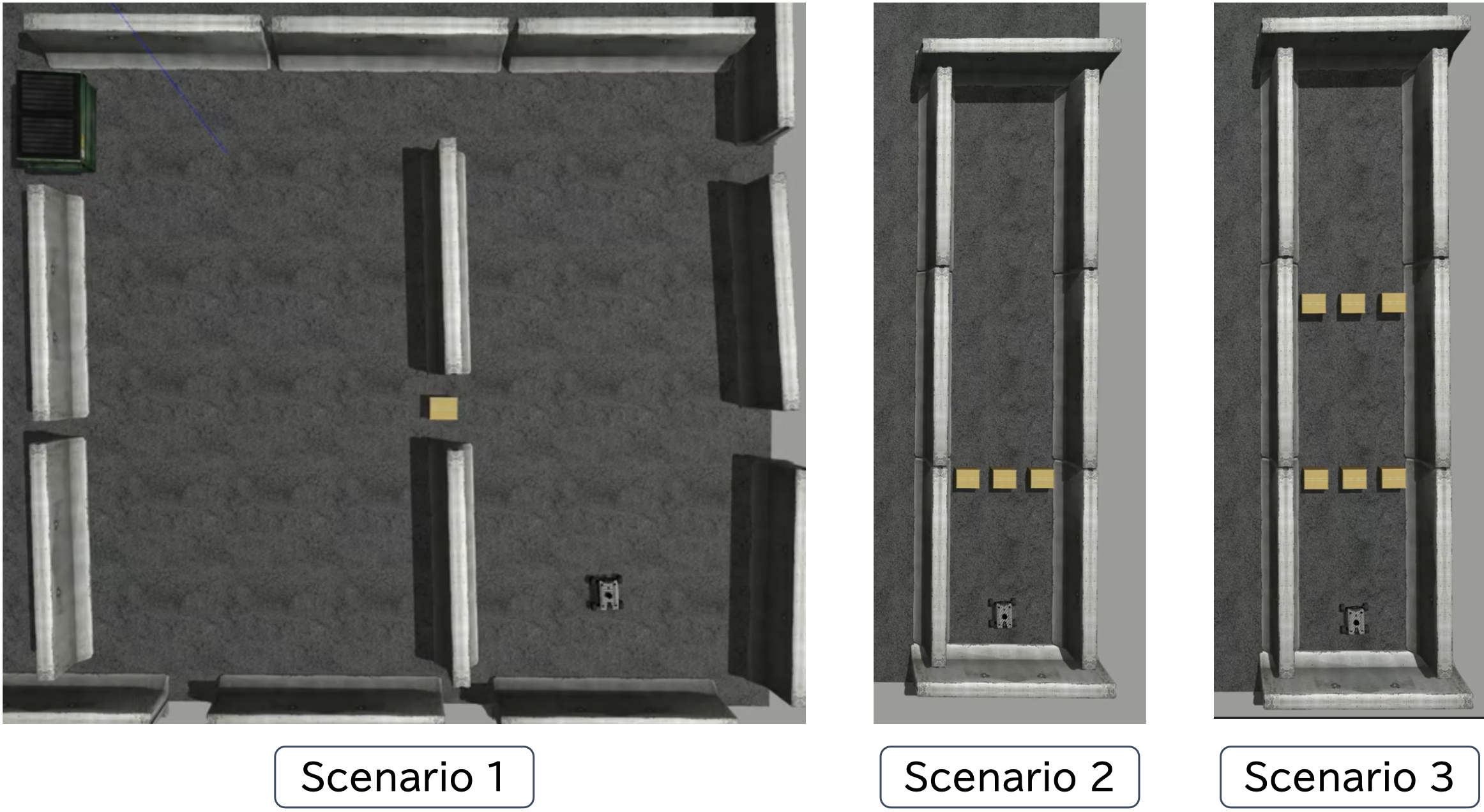}
\caption{The maps used as the respective 3 scenarios for the experimental evaluation.}
\label{Fig:worlds}
\end{figure}

\section{Experimental Evaluation}

This section describes the simulation experiments conducted to validate the performance and effectiveness of the proposed adaptive cost-map-based robot path planning system.

The experiments evaluate the robot's navigation capabilities and obstacle movability estimation accuracy within various scenarios designed in Gazebo. The metrics selected for assessment clearly demonstrate the system's adaptability and efficiency compared to conventional static mapping methods.

\subsection{Simulation Scenarios}

The simulation tests were conducted across three primary scenarios in customized Gazebo environments (see Fig. \ref{Fig:worlds}), specifically created to challenge different aspects of our adaptive cost-map-based navigation system.

Scenario 1 (Individual Obstacle Interactions): This scenario involved an isolated obstacle with different movability levels: light (1-a), heavy (1-b), and immovable (1-c). This setup provided a clear baseline for evaluating the adaptive layer's ability to initially classify and adjust obstacle costs based on real-time interactions. 

Scenario 2 (Corridor Mixed Configurations): The corridor setups combined multiple obstacles with varied movability levels, arranged to necessitate sequential interactions. In configurations 2-a, 2-b, and 2-c, the central obstacle directly in the robot's path had differing movability: light, heavy, and immovable respectively. These setups simulated realistic constrained environments such as hallways cluttered with mixed objects.

Scenario 3 (Complex Double-row Obstacles): Designed to test the system's robustness in more complicated situations and recovery capabilities, this scenario involved parallel rows of obstacles with identical movability properties. The complexity of navigating and interacting sequentially to clear a feasible path provided a rigorous test of our adaptive mechanism.

\begin{figure}[h]
\centering
\includegraphics[width=0.48\textwidth]{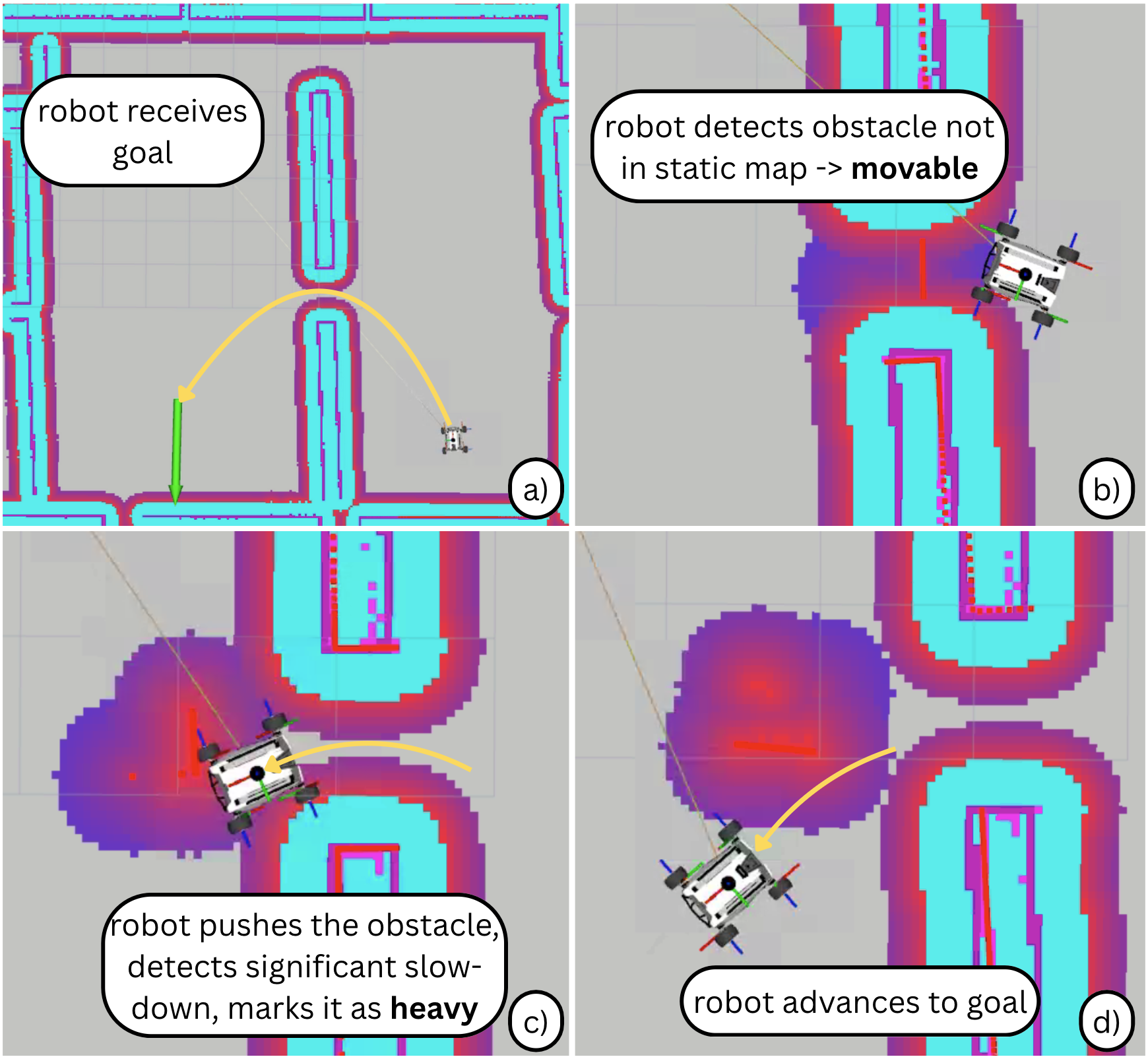}
\caption{Scenario 1-b (heavy obstacle): annotated costmap sequence showing the initial plan, insertion of an unmapped return as “movable,” slowdown-triggered escalation to “heavy,” and the resulting re-route to the goal.}
\label{Fig:scenario1b}
\end{figure}

\subsection{Performance Metrics}

To quantitatively assess the navigation capabilities, we established the following metrics:

\textit{Goal Success Rate:} Defined as the percentage of successful navigation attempts to the goal out of 20 trials per scenario.

\textit{Navigation Time:} Comparison of traversal times required for the robot to reach the goal, specifically contrasting performance with and without the adaptive cost-map layer enabled. Shorter navigation times indicate improved efficiency.

\textit{Movability Estimation Accuracy:} Evaluated based on the correctness of the adaptive layer's real-time obstacle classification compared to ground truth set in simulation (light, heavy, immovable).

\begin{figure}[h]
\centering
\includegraphics[width=0.48\textwidth]{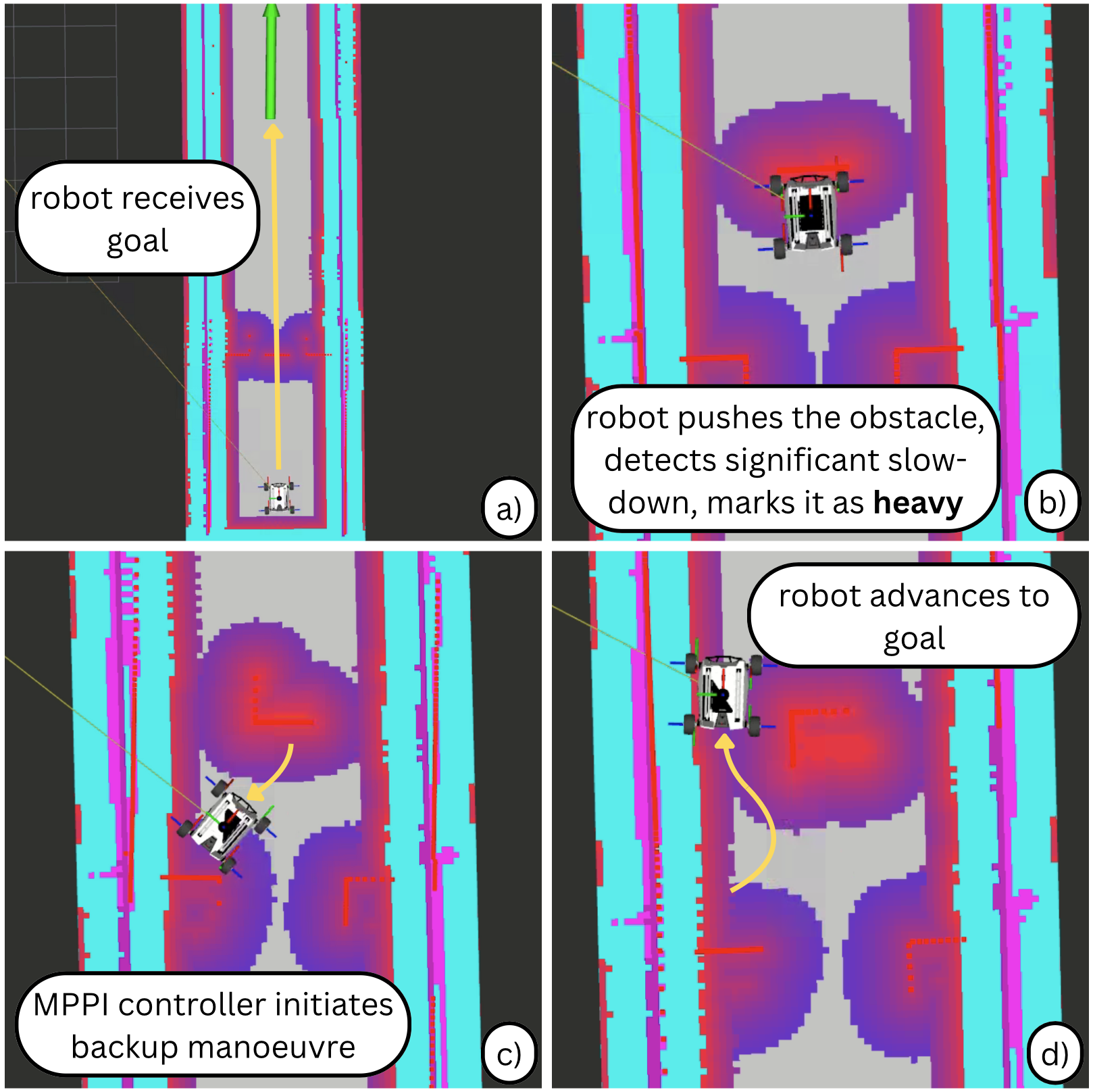}
\caption{Scenario 2-b (corridor, heavy center obstacle): annotated costmap sequence (a–d) showing the initial plan; there was slowdown on contact so the robot escalated the cost to heavy; MPPI triggers a backing maneuver; global plan re-routes and the robot proceeds to the goal.}
\label{Fig:scenario2b}
\end{figure}

\subsection{Results and Discussion}

The results obtained from the simulations (see Table 1) are summarized and discussed as follows:

Scenario 1 illustrates the adaptive layer’s capacity to estimate obstacle movability through direct interaction. In 1-a, the robot successfully identified and navigated past light obstacles with minimal difficulty. Scenario 1-b (see Fig. \ref{Fig:scenario1b} confirmed that while heavy obstacles were still occasionally pushable, their interaction results varied depending on contact angle and robot speed, leading to reduced success rates and contributing to the drop in movability estimation accuracy. Scenario 1-c introduced an immovable obstacle; the robot initially attempted to push it, detected failure via progress checking, and rerouted accordingly. This demonstrates a deliberate trade-off: longer traversal time in exchange for learning and avoiding repeated futile interactions.

\begin{table}[!t]
\centering
\caption{Navigation Results across Scenarios}
\label{tab:results}
\begin{tabular}{|c|c|c|c|}
\hline
\textbf{Scenario} & \shortstack{\textbf{Goal}\\\textbf{Success Rate}} & \shortstack{\textbf{Time}\\\textbf{(adaptive vs. no layer)}} & \shortstack{\textbf{Movability}\\\textbf{Accuracy}} \\
\hline
1-a & 100\% & 27.79s vs. 41.71s (67\%) & 100\% \\
1-b & 80\% & 34.24s vs. 41.71s (82\%) & 70\% \\
1-c & 85\% & 60.18s vs. 41.71s (144\%) & 95\% \\
2-a & 100\% & impossible & 95\% \\
2-b & 100\% & impossible & 60\% \\
2-c & 95\%  & impossible & 80\% \\
3   & 100\% & impossible & 80\% \\
\hline
\end{tabular}
\end{table}

\begin{figure}[b!]
\centering
\includegraphics[width=0.48\textwidth]{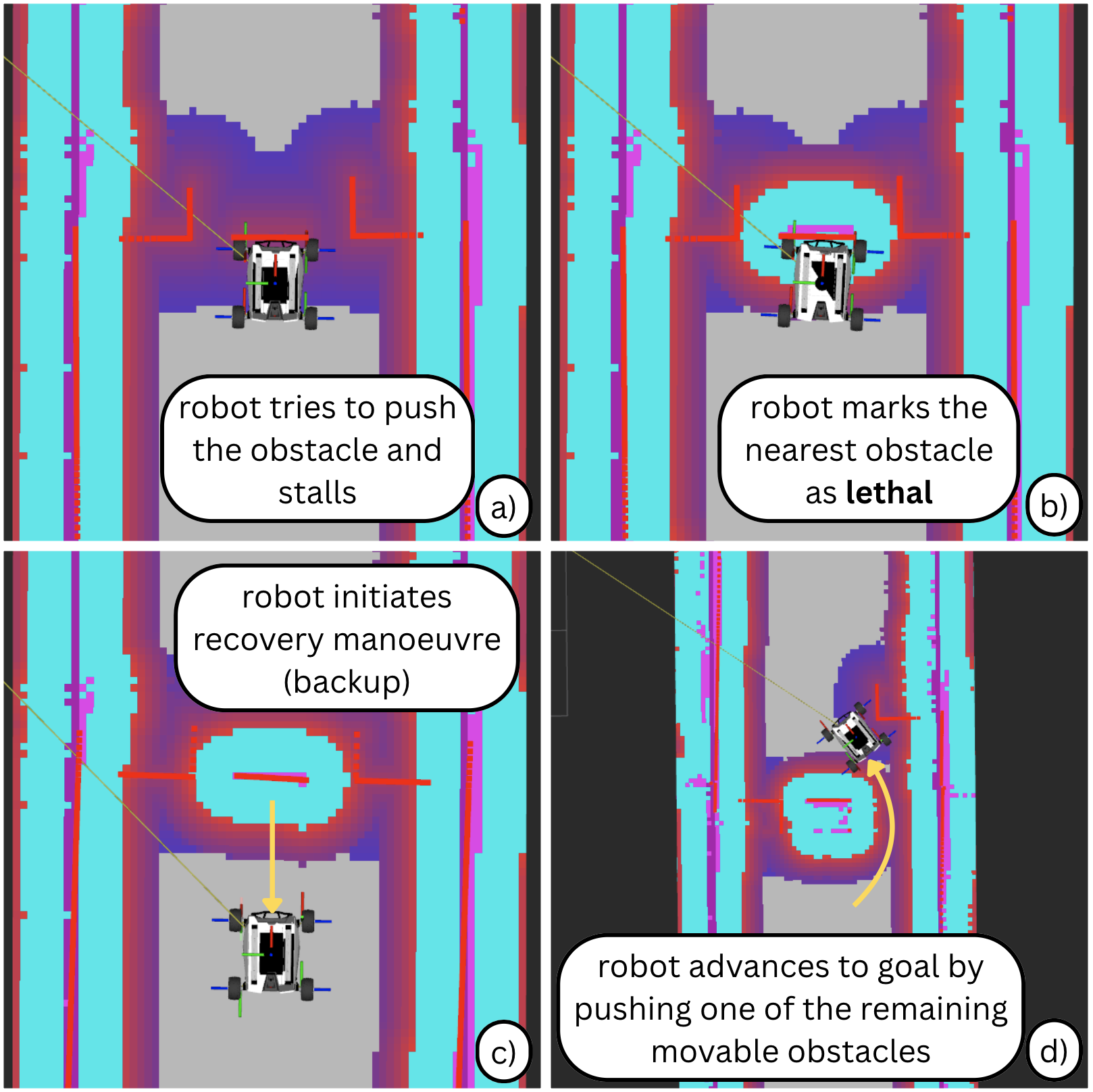}
\caption{Scenario 2-c (corridor, immovable center obstacle): annotated costmap sequence (a–d) showing attempted push and stall; escalation of the nearest obstacle to lethal; Recovery behavior (back-up); re-route to the goal by pushing a remaining movable side obstacle.}
\label{Fig:scenario2c}
\end{figure}

Scenario 2 emphasizes the necessity of adaptive movability classification. Without the adaptive layer, baseline methods failed entirely due to treating all obstacles as static and unmovable, with no mechanism to revise that classification. The adaptive system, by contrast, succeeded across all corridor configurations. Notably, light obstacles remained easy to push and were consistently classified correctly at the end. However, for heavy obstacles (scenario 2-b, see Fig. \ref{Fig:scenario2b}), estimation accuracy again dropped, highlighting the challenge of distinguishing pushable-but-difficult objects under varying approach conditions. Figure \ref{Fig:scenario2c} shows how the robot would act in scenario 2-c.

Scenario 3 was designed to test the system in a more complex and cluttered environment. The robot successfully navigated through the double-row setup where all obstacles were light. However, the tight spaces and increased number of obstacles led to clustering and localization difficulties, particularly affecting consistent obstacle distinction and map alignment.

These issues in Scenario 3 reflect a broader pattern observed across all scenarios. First, clustered obstacles, particularly those near walls, occasionally led to ambiguous or erroneous classification due to occlusion in LiDAR data. Second, localization drift caused misalignments in costmap updates, resulting in inconsistent obstacle treatment. Finally, while the velocity-based progress checker generally performed well, it sometimes struggled to distinguish between legitimate low-speed maneuvering and actual pushing failure, especially during tight turns or cautious movements.


\section{Conclusions and Future Work}
This research introduced an adaptive cost-map-based path-planning system designed for autonomous robot navigation in unknown environments cluttered with movable obstacles. The system assigns a provisional ‘movable’ cost to any LiDAR returns not in the static map and updates that cost online when interaction cues (slowdown/stall) indicate resistance. Core contributions include the development of the specialized "Movable Obstacles Layer" plugin for ROS2’s Nav2 costmap stack, an adaptive costmap methodology that dynamically adjusts obstacle traversal costs, and a velocity-based progress checker that monitors interaction feedback to reassess obstacle movability during navigation. Simulations across seven Gazebo scenarios indicate improved goal-reach rates and fewer deadlocks relative to a static obstacles cost-map baseline that cannot push; traversal times were mostly improved, and movability cues were reliable for light/immovable objects but less so for heavy ones.

The primary strengths of our approach include its generalizability across diverse obstacle setups, the simplicity and computational efficiency of integration into standard ROS2 navigation frameworks, and the elimination of the need for extensive pre-trained semantic or vision-based models.

Despite these strengths, several challenges remain. Using robot velocity alone to infer pushing effectiveness revealed limitations, particularly in instances of near-stalling or during turns. Additionally, occasional localization drift led to temporary costmap inaccuracies, affecting obstacle status updates. Obstacle clustering also posed challenges, occasionally resulting in ambiguous classifications due to partial sensor occlusion.

Addressing these limitations offers clear avenues for future research. We plan to enhance the reliability of movability estimation by incorporating motor effort and load measurements, providing a complementary metric alongside velocity-based methods. Advanced clustering and obstacle disambiguation algorithms will also be explored to better distinguish obstacles in close proximity to static structures. Further improvements in localization and map consistency techniques will be integrated to reduce transient costmap errors. Finally, transitioning from simulation-based evaluations to real-world robotic deployments in indoor environments will be pursued to comprehensively assess the practical applicability and robustness of our adaptive navigation approach.





\bibliographystyle{IEEEtran}
\bibliography{IEEEabrv,root}

%
%
%

\end{document}